%% file: main.tex
\documentclass{article}

\PassOptionsToPackage{numbers, compress}{natbib}

\usepackage[preprint]{neurips_2024}             

\usepackage[utf8]{inputenc}
\usepackage[T1]{fontenc}
\usepackage{url}
\usepackage{booktabs}
\usepackage{amsfonts}
\usepackage{nicefrac}
\usepackage{microtype}
\usepackage{xcolor}
\usepackage{times}
\usepackage{epsfig}
\usepackage{amsmath}
\usepackage{makecell}
\usepackage{tabularx}
\usepackage{diagbox}
\usepackage{enumitem}
\usepackage{xspace}
\usepackage{graphicx}
\usepackage{subcaption}
\usepackage{multirow}
\usepackage{colortbl}
\usepackage{pifont}
\usepackage{wrapfig}
\usepackage{tcolorbox}
\usepackage{listings}
\usepackage{float}
\usepackage{marvosym}
\usepackage{CJKutf8}
\usepackage{array}

\definecolor{citecolor}{HTML}{0071BC}
\usepackage[
    pagebackref,
    breaklinks=true,
    colorlinks=true,
    linkcolor=red,
    citecolor=citecolor,
    urlcolor=black,
    bookmarks=false
]{hyperref}

\makeatletter
\def\@maketitle{%
  \vbox{%
    \hsize\textwidth
    \linewidth\hsize
    \vskip 0.1in
    \noindent
    \includegraphics[height=0.24in]{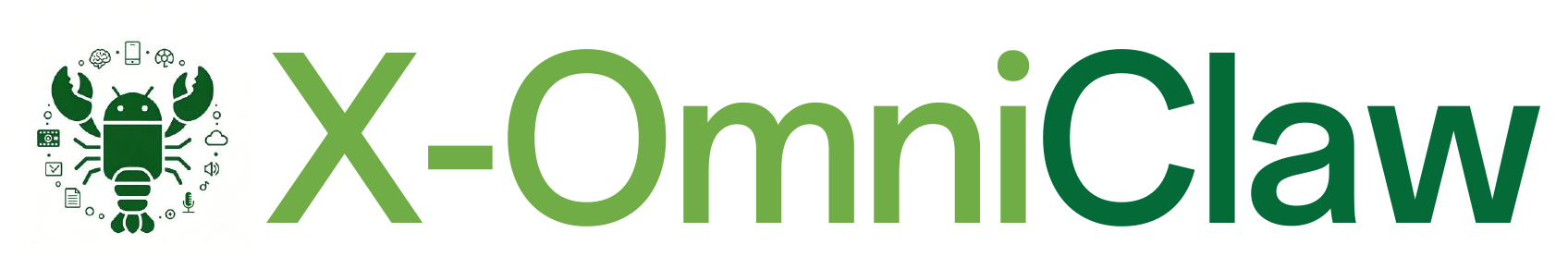}%
    \vskip 0.22em
    \@toptitlebar
    \centering
    {\LARGE\bf \@title\par}%
    \@bottomtitlebar
    \if@submission
      \begin{tabular}[t]{c}\bf\rule{\z@}{24\p@}
        Anonymous Author(s) \\
        Affiliation \\
        Address \\
        \texttt{email} \\
      \end{tabular}%
    \else
      \def\And{%
        \end{tabular}\hfil\linebreak[0]\hfil%
        \begin{tabular}[t]{c}\bf\rule{\z@}{24\p@}\ignorespaces%
      }%
      \def\AND{%
        \end{tabular}\hfil\linebreak[4]\hfil%
        \begin{tabular}[t]{c}\bf\rule{\z@}{24\p@}\ignorespaces%
      }%
      \begin{tabular}[t]{c}\bf\rule{\z@}{24\p@}\@author\end{tabular}%
    \fi
    \vskip 0.3in \@minus 0.1in
  }%
}
\makeatother

\newcommand{\eg}{{\emph{e.g.}}}

\renewcommand{\ghlink}{https://github.com/OPPO-Mente-Lab/X-OmniClaw}

\title{X-OmniClaw Technical Report: A Unified Mobile Agent for Multimodal Understanding and Interaction}

\author{%
\textbf{Xiaoming Ren, Ru Zhen, Chao Li, Yang Song, Qiuxia Hou, Yanhao Zhang$^{\dagger}$$^{\ast}$, Peng Liu,}\\
\textbf{Qi Qi, Quanlong Zheng, Qi Wu, Zhenyi Liao, Binqiang Pan, Haobo Ji, Haonan Lu$^{\ast}$}\\[0.25em]
{\footnotesize\texttt{\{renxiaoming1,zhenru1,zhangyanhao,luhaonan\}@oppo.com}}\\
\\[-0.35em]
\textbf{Multi-X Team, OPPO AI Center}\\
\\[-0.35em]
\github ~ \url{\ghlink}
}

\begin{document}
\begin{CJK}{UTF8}{gbsn}

\maketitle
\begingroup
\renewcommand{\thefootnote}{}
\footnotetext{$^{\dagger}$~Project Leader $^{\ast}$~Corresponding Author.}
\endgroup


\begin{abstract}
    Inspired by the development of OpenClaw, there is a growing demand for mobile-based personal agents capable of handling complex and intuitive interactions. In this technical report, we introduce \textbf{X-OmniClaw}, a unified mobile agent designed for multimodal understanding and interaction in the Android ecosystem. This unified architecture of perception, memory, and action enables the agent to handle complex mobile tasks with high contextual awareness. Specifically, \textbf{Omni Perception} provides a unified multimodal ingress pipeline that integrates UI states, real-world visual contexts, and speech inputs, leveraging a temporal alignment module to decompose raw data into structured multimodal intent representations. \textbf{Omni Memory} leverages multimodal memory optimization to enhance personalized intelligence by integrating runtime working memory for task continuity with long-term personal memory distilled from local data, enabling highly context-aware and personalized interactions. Finally, \textbf{Omni Action} employs a hybrid grounding strategy that combines structural XML metadata with visual perception for robust interaction. Through Behavior Cloning and Trajectory Replay, the system captures user navigation as reusable skills, enabling precise direct-access execution. Demonstrations across diverse scenarios show that \textbf{X-OmniClaw} effectively enhances interaction efficiency and task reliability, providing a practical architectural blueprint for the next generation of mobile-native personal assistants.
    \end{abstract}
    

\input{sec/1_introduction}

\input{sec/2_architecture}

\input{sec/3_omni_perception}

\input{sec/4_omni_memory}
\input{sec/5_omni_action}

\input{sec/6_use_cases_and_demo_scenarios}
\input{subsec/6_1}

\input{subsec/6_2}
\input{subsec/6_3}

\input{sec/7_conclusion}

{
    \small
    \bibliographystyle{plain}
    \bibliography{main}
}

\end{CJK}
\end{document}

%% file: sec/1_introduction.tex
\section{Introduction and Related Work}
\label{sec:intro_related}

\subsection{Introduction}

The evolution of Large Language Models (LLMs) is moving beyond semantic dialogue toward MLLM-based agents capable of autonomous task execution. In this context, the smartphone---acting as a high-frequency extension of human activity---functions as a digital sensory organ, providing a versatile foundation for the perception-heavy requirements of mobile agents. While solutions like Doubao Phone have verified the engineering feasibility of cross-app orchestration on Android~\cite{qin2025uitars}, they often lack deep control for user-defined logic and customization. In contrast, the rapid adoption of OpenClaw~\cite{openclaw2026repo} highlights a strong demand for localized, user-steerable execution frameworks. However, OpenClaw remains centered on PC-side execution, which is detached from the dynamic mobile contexts required for real-time interaction.

The design of \textbf{X-OmniClaw} aims to bridge the gap between architectural execution and mobile-native autonomy. Here, \textit{Omni} denotes the integration of three sensing domains: on-screen UI state, real-world visual context, and audio input. The prefix \textit{X} further emphasizes the cross-modal nature of the system, evolving it into a unified perception-to-action framework for reliable task execution.

Inspired by OpenClaw, \textbf{X-OmniClaw} leverages the smartphone as a persistent interface for multimodal data streams. By integrating environmental sensing with application-level control, the framework enables the agent to utilize real-world context when executing digital tasks. This approach establishes the mobile agent as a functional automation tool, providing a streamlined solution for complex task completion across diverse, mobile-centric environments. The collaboration between Omni Perception, Omni Memory, and Omni Action enables the agent to process richer environmental data, maintain task continuity over time, and execute complex operations more effectively.

\subsection{Related Work}

\paragraph{Open-source frameworks.}
OpenClaw represents an important open-source direction for agent engineering by placing a layered control system around the model~\cite{openclaw2026repo}. Its architecture decouples the model layer, core runtime, skills, and external interfaces~\cite{yao2022react}, turning agent capability into explicit behavioral rules, persistent storage, and atomic tool abstractions. The key idea is that structured skills can reduce the randomness of model outputs~\cite{wang2023voyager}, while persistent memory helps maintain logical consistency across long-horizon workflows~\cite{openclaw2026memory}.
Hermes Agent, developed by Nous Research, offers a complementary "learning-first" paradigm in agent architecture design~\cite{hermesagent2026repo}. Its core innovation lies in a self-improving learning loop that autonomously generates and refines reusable procedural skills from interaction data, combined with a three-tier memory hierarchy (short-term inference memory, procedural skill documents, and contextual persistence) that mimics human procedural learning~\cite{hermesagent2026docs}. Unlike OpenClaw's explicit control via structured skills, Hermes emphasizes emergent capability growth through automated skill creation while maintaining compatibility with standard agent tool ecosystems.
This dual approach—externalized control logic (OpenClaw) and autonomous capability evolution (Hermes Agent)—improves execution determinism and also gives users substantial freedom to customize, extend, and redesign the agent's operating logic, addressing both reliability and adaptability needs in complex agent deployments~\cite{openclawhermes2026comparison}.

\paragraph{Mobile perception, execution and simulation-based agents.}
Research on mobile agents has explored how an agent can perceive and operate app interfaces under dynamic GUI conditions~\cite{rawles2023aitw}. Mobile-Agent~\cite{wang2024mobileagent} and AppAgent~\cite{han2023appagent} investigate the feasibility of purely visual interaction, where the agent relies on screenshots and coordinate-level grounding~\cite{wu2024osatlas} to locate interface elements and perform actions. Industrial systems such as Doubao Phone further demonstrate that mobile automation can be scaled through the combination of visual foundation models and system-level orchestration engines, a trend also exemplified by UI-TARS~\cite{qin2025uitars}. Parallel to real-world agent design, another line of work studies mobile decision making via simulated environments and reinforcement learning. Platforms such as AndroidWorld~\cite{rawles2025androidworld}, OSWorld~\cite{xie2024osworld}, and WebArena~\cite{zhou2023webarena} provide controlled testbeds for repeated interaction and evaluation, while methods such as DigiRL~\cite{bai2024digirl} explore iterative optimization to enhance action stability under dynamic and partially observable UI states. Collectively, these studies validate the feasibility of mobile task execution and advance policy robustness in constrained settings, yet they still struggle to guarantee controllability and transparency in real-world deployment, and pay limited attention to end-user governance and customizable reshaping of the underlying execution framework.

\paragraph{Cloud-Centric vs. Edge-Native Architectures}
Existing mobile agent frameworks follow a \textbf{cloud-centric} paradigm. This approach operates by running virtualized Android instances in remote data centers, as exemplified by platforms such as RedFinger~\cite{redfinger}, Wuying~\cite{wuying}, and Tencent Cloud Phone~\cite{tencentcloudphone}. In these systems, the agent operates within a simulated environment detached from the physical entity. While this reduces the demand for local computational power, it inherently lacks access to the user's authentic local hardware (e.g., sensors, local cameras), system-level configurations, and private local data. Furthermore, it imposes the burden of maintaining a separate cloud identity. In contrast, \textbf{X-OmniClaw} introduces an \textbf{edge-native} architecture that executes directly on the user's physical device, thereby eliminating the gap between simulated environments and real-world interaction contexts.

\textbf{X-OmniClaw} emerged in the broader wave of developer interest in open mobile automation enlightened by OpenClaw~\cite{openclaw2026repo}. Our project was inspired by AndroidClaw ~\cite{hermesapp2026repo} ~\cite{operit2026repo}, and have since built a set of distinctive core capabilities on top of this baseline, as elaborated in the following sections.

%% file: sec/2_architecture.tex
\section{Frameworks of X-OmniClaw}
\label{sec:architecture}

\begin{figure}[htbp]
  \centering
  \includegraphics[width=\linewidth]{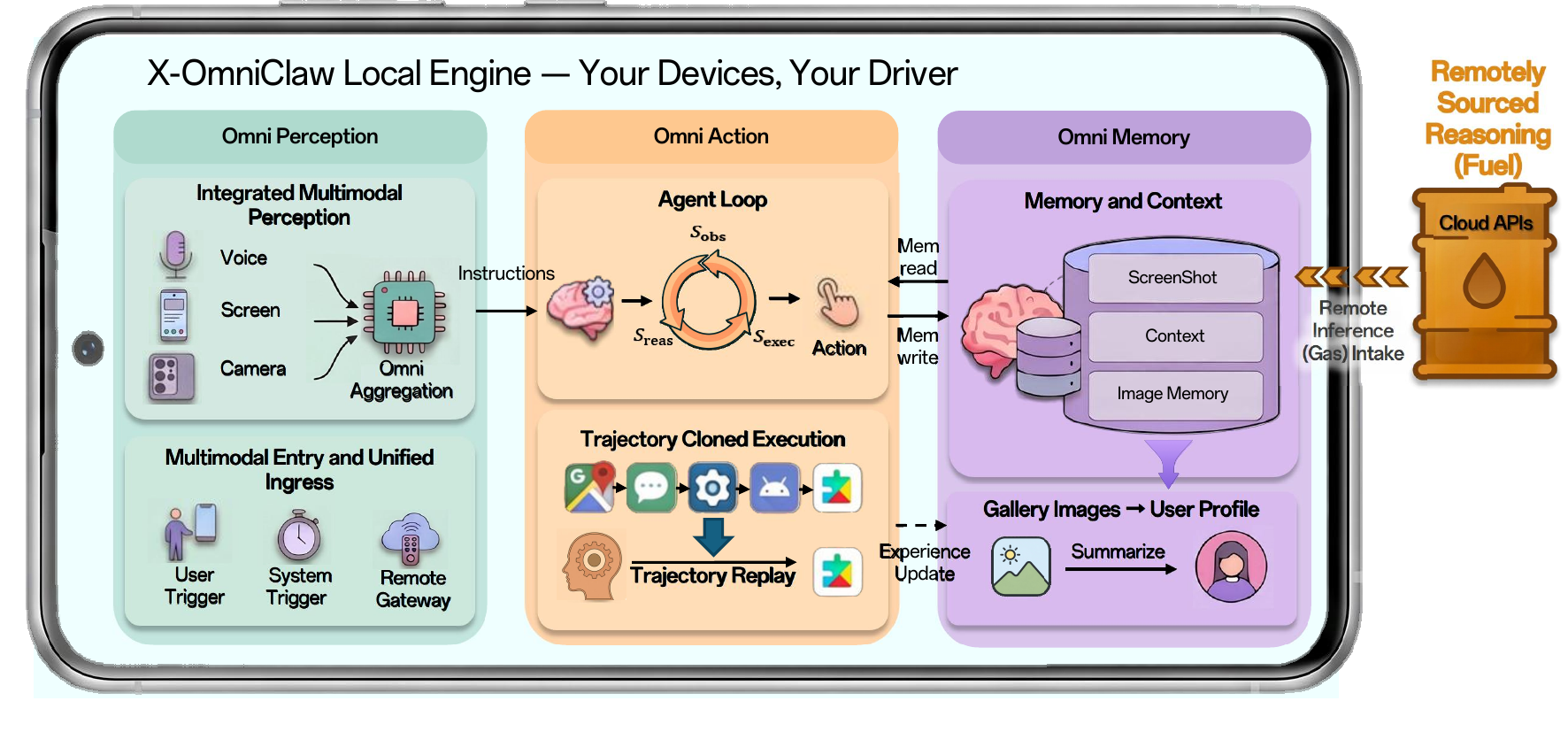}
  \caption{Overall system structure of X-OmniClaw.}
  \label{fig:x-omniclaw-structure}
\end{figure}

\textbf{X-OmniClaw} targets Android mobile-agent settings where the assistant must sustain continuous perception while still executing device actions reliably. This section gives a system-level view of the framework: we argue that perception, memory, and action are not independent modules to bolt on, but a single co-designed stack. Figure~\ref{fig:x-omniclaw-structure} summarizes the concrete architecture---integrated multimodal perception (Voice, Screen, and Camera) drives on-device execution via the agent loop, which is then transformed into refined experience and persistent memory to iteratively optimize future performance. The following subsections unpack these components in greater detail.

\subsection{Edge-Native Architectures}
Departing from the aforementioned cloud-centric architectures, the core logic of \textbf{X-OmniClaw} resides entirely on the user's local Android device. To use a car analogy: the smartphone serves as the vehicle, \textbf{X-OmniClaw} acts as the internal engine for control and perception, while the cloud-based LLM functions solely as the ``fuel'' for high-level reasoning. By deploying core perception and execution capabilities locally, the system receives on-demand computational support from the cloud, eliminating the need to host heavy inference models on the smartphone itself.

This design enables the agent to directly manipulate authentic applications and system settings without the extra burden of maintaining a cloud phone. Operationally, the system follows a compact execution pipeline: multimodal triggers are first captured from user input and device context, then interpreted by a central planning process, and finally grounded into concrete Android operations through reusable skills and tool interfaces. These components converge into three core functional modules---Omni Perception, Omni Memory, and Omni Action---forming a tightly coupled stack for edge-native mobile agency.

\subsection{Overview of Core Capabilities}
Based on this architecture, we present \textbf{X-OmniClaw}, an omni-modal mobile agent that unifies smartphone interaction across three pillars:

\begin{itemize}
    \item \textbf{Omni Perception} serves as the system's multimodal ingress, integrating UI states, real-world visual contexts, and speech inputs. It decomposes raw streaming data into structured intents, which then drive the subsequent reasoning and execution loops.
    \item \textbf{Omni Memory} maintains task continuity by unifying runtime working memory with long-term personal knowledge. It continuously updates the user profile with semantic insights distilled from device-resident personal data, providing the persistent context required for personalized, multi-turn interactions.
    \item \textbf{Omni Action} implements a robust execution framework that combines structural XML and visual information. Through behavior cloning and trajectory replay, it transforms complex user navigation into reusable skill trajectories, enabling the agent to translate high-level intent into precise, hardware-level actions while maintaining state consistency through continuous interaction with the Memory module.
\end{itemize}

Together, these components enable a mobile agent that can perceive richer context, preserve continuity over time, and execute complex real-world tasks more reliably.

%% file: sec/3_omni_perception.tex
\section{Omni Perception}
\label{sec:cross-world-perception}
\begin{figure}[htbp]
    \centering
    \includegraphics[width=\linewidth]{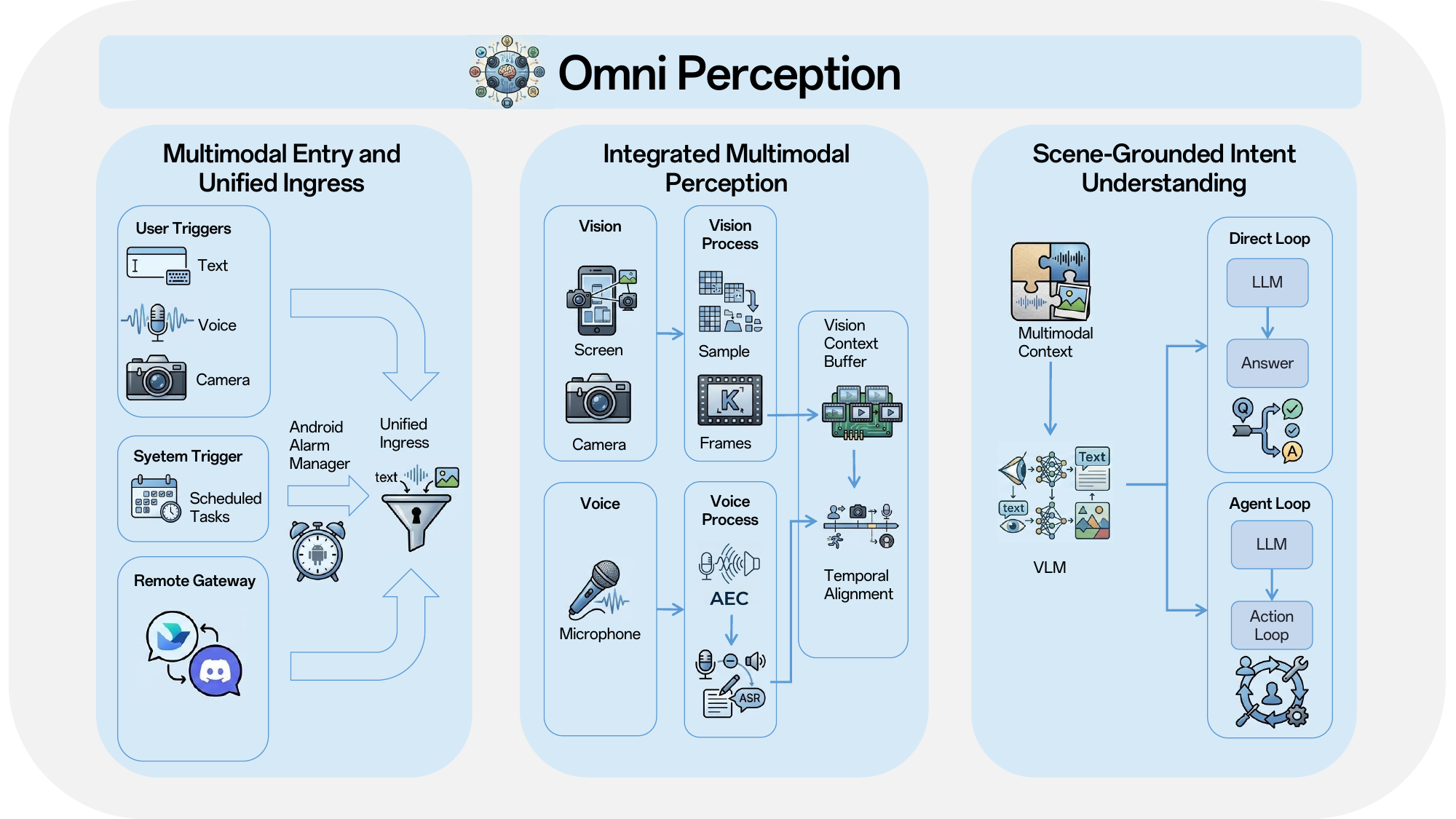}
    \caption{Overview of Omni Perception: multimodal entry, multimodal perception, and scene-grounded intent understanding.}
    \label{fig:omni-perception-overview}
  \end{figure}
\paragraph{Multimodal Entry and Unified Ingress.}
X-OmniClaw establishes a unified gateway to consolidate diverse multimodal inputs. Requests may originate from direct user triggers, such as in-app UI interactions, system-level floating widgets, and microphone input, from user-defined proactive triggers such as scheduled tasks, or from external ecosystems such as Feishu, Discord bots, and other remote gateways. All of these requests are funneled into the same system pipeline. For recurring on-device tasks, we additionally use Android \texttt{AlarmManager} to build a system-level wake-up path. This allows the system to receive scheduled and repeated triggers even under standby or low-power conditions, and to merge them back into the same unified entry point with semantics consistent with immediate interaction.


\paragraph{Integrated Multimodal Perception.}
X-OmniClaw combines the sensing channels available on the phone into a first-person multimodal perception system that jointly models on-screen UI state, real-world visual context, and audio input. Camera streams and screen projection capture the visual environment across both domains, while speech recognition transcribes microphone input in real time. To handle the common mobile case in which the device is simultaneously playing audio, the system further applies on-device adaptive acoustic echo cancellation (AEC) to suppress self-generated interference during collection. At the implementation level, these signals are organized through a decoupled streaming pipeline: visual observations are pushed asynchronously into an in-memory ring buffer that preserves short-term history, and a temporal alignment module matches speech and visual streams through shared timestamps.


\paragraph{Scene-Grounded Intent Understanding.}
When multimodal input enters the system, X-OmniClaw does not immediately trigger downstream actions. Instead, a VLM first interprets the current visual scene together with the user's query and expands the raw input into a more complete semantic representation of intent. If the user's question can be answered directly from the current scene, the system returns an answer immediately. Otherwise, the decomposed result is converted into a structured intent representation and passed to the downstream AgentLoop for execution. For example, when a user asks, "How much does this cost on Taobao?", the system may first infer from the visual context that the referenced object is an Evian spray, reformulate the request as "the user wants to know the price of Evian spray on Taobao," and only then launch Taobao for the subsequent search and interaction.

%% file: sec/4_omni_memory.tex
\section{Omni Memory}
\label{sec:multimodal-long-term-memory}

\paragraph{Working Memory and Long-Term User Memory.}
To build long-term memory, the system first has to keep track of what is happening in the current task. In practice, this means preserving a multimodal runtime context across multiple turns, foreground changes, and app switches. That context is not just a text history: it includes screenshots as visual evidence, compressed observations as distilled semantic context, and execution state as a record of task progress. Together, these signals act as the agent's working memory. They allow the system to resume a task without losing its place, relate new observations to earlier evidence, and maintain a trace of what has already happened. This runtime continuity is what lets X-OmniClaw operate as an ongoing device agent rather than a one-shot response system.


While this working memory ensures runtime continuity, Omni Memory further extends the agent's capability by distilling long-term multimodal context from local personal data. The system distills multimodal information from the user's local data environment---including personal media assets, interaction trajectories, and task-relevant metadata---into persistent memory artifacts and user-profile representations. These multimodal memories can be injected into downstream reasoning and interaction contexts, enabling the agent to provide more personalized responses, preserve cross-modal context across tasks, and avoid repeatedly reconstructing user-specific information from scratch.

A concrete example is the user's photo gallery. Instead of relying only on raw images, X-OmniClaw transforms visual assets such as gallery photos into compact, structured semantic records that capture objects, scenes, events, and user-relevant cues. These records support image-grounded question answering, semantic retrieval over past photos, and personalized media selection for later automation workflows.

\begin{figure}[htbp]
  \centering
  \includegraphics[width=\linewidth]{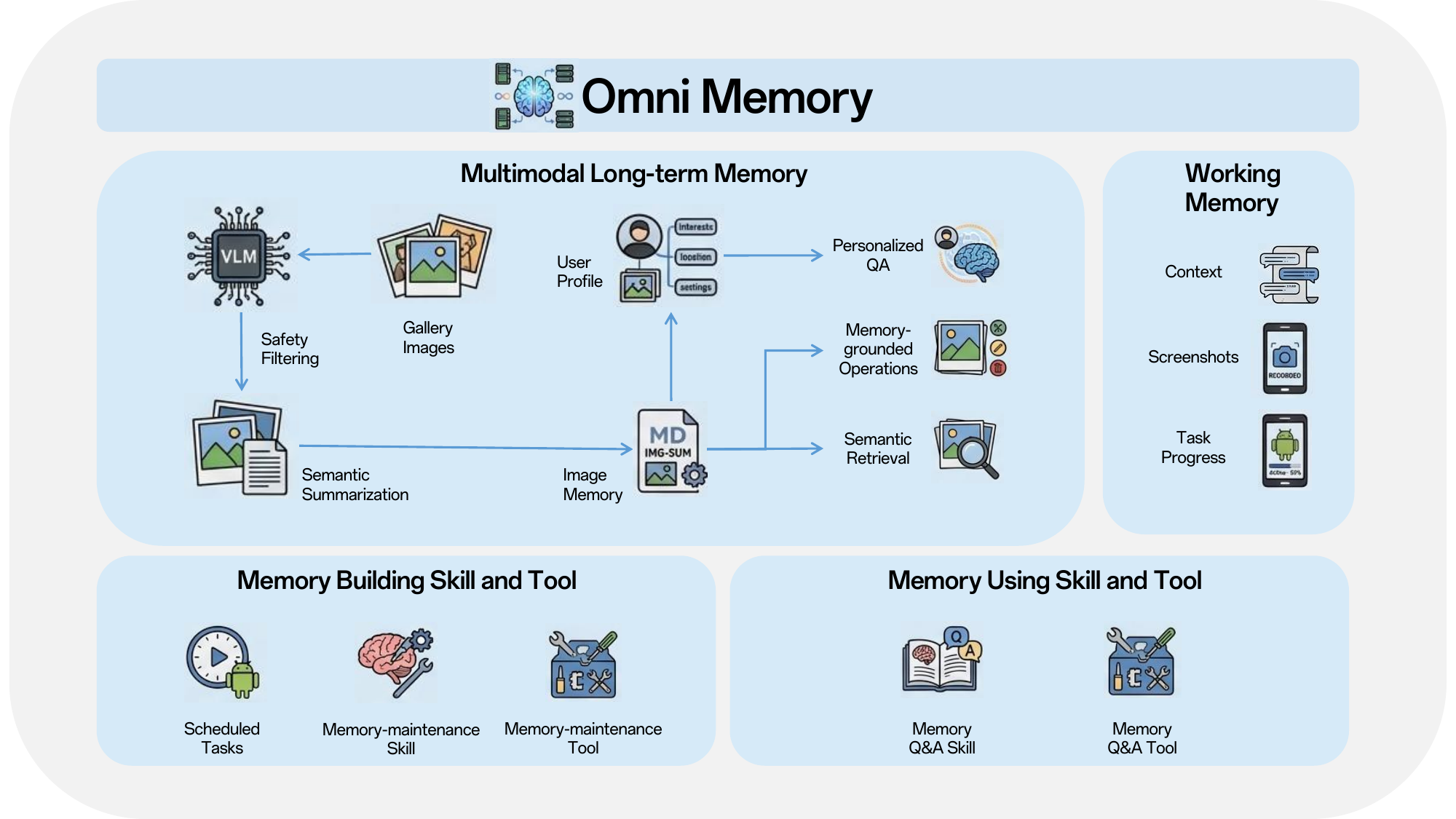}
  \caption{Overview of Omni Memory: runtime context, long-term artifacts, and Skill--Tool coordination.}
  \label{fig:omni-memory-overview}
\end{figure}

\paragraph{How Memory Is Built, Used, and Secured.}
In practice, this capability is implemented through \textbf{Skill}--\textbf{Tool} coordination. Skills define the workflow and division of labor: some are responsible for memory maintenance, such as synchronization, update, and rebuild, while others are responsible for memory consumption, such as question answering, retrieval, and memory-grounded operations. Tools execute the concrete steps that make these workflows actionable. During image processing, the system prioritizes multimodal models for semantic summarization; if model invocation fails, it falls back to simplified summaries derived from image metadata so that the pipeline can continue rather than break. 

More broadly, memory production is separated from memory consumption, which reduces workflow entanglement and makes the system easier to iterate and stabilize over time. Before anything is written into memory, the system applies a unified filtering and redaction step. The goal is to reduce the chance that sensitive information is stored in long-term memory. The user is also given explicit controls over whether gallery memory is enabled and whether the derived user profile is injected into downstream context. To reduce the upload risk associated with cloud vision, a natural next step is to move semantic image summarization onto on-device models so that raw pixels stay on the device as much as possible.

%% file: sec/5_omni_action.tex
\section{Omni Action}
\label{sec:skill-clone}

\subsection{Omni Action in the App Ecosystem}

Android applications are highly heterogeneous in rendering style, interface exposure, and interaction logic, so mobile execution cannot rely on a single source of interface evidence. To handle this complexity, X-OmniClaw adopts a dynamic strategy that leverages its visual understanding to balance structural and visual evidence across interfaces.

X-OmniClaw organizes each action as a loop of observation, reasoning, and execution. During observation, we build a unified observation stack from multimodal interface evidence. The agent loop then reasons over this stack to observe the current page and understand the status of the previous step, select the appropriate skill, retrieve relevant memory when needed, and return either the next action or a direct response. The resulting decision is finally executed through a diverse set of action modalities: these include not only Android-level atomic operations, but also higher-level operations such as file-system manipulations, RAG and other predefined tools. The key to the observation stage is \textbf{hybrid UI understanding}: the system combines XML signals, an on-device grounding model, and OCR to localize actionable targets with higher precision. Structured interface information is used when it is reliable, while visual grounding and text recognition compensate when structural cues are weak, incomplete, or spatially ambiguous. This mechanism is especially effective in advertisement-heavy or visually cluttered interfaces, where XML alone may not provide a precise click location. In such cases, visual information supplements the missing spatial evidence and helps the system execute more accurate clicks. By integrating the omni perception capabilities described above, this dynamic strategy improves end-to-end action robustness and accuracy.

\begin{figure}[htbp]
  \centering
  \includegraphics[width=\linewidth]{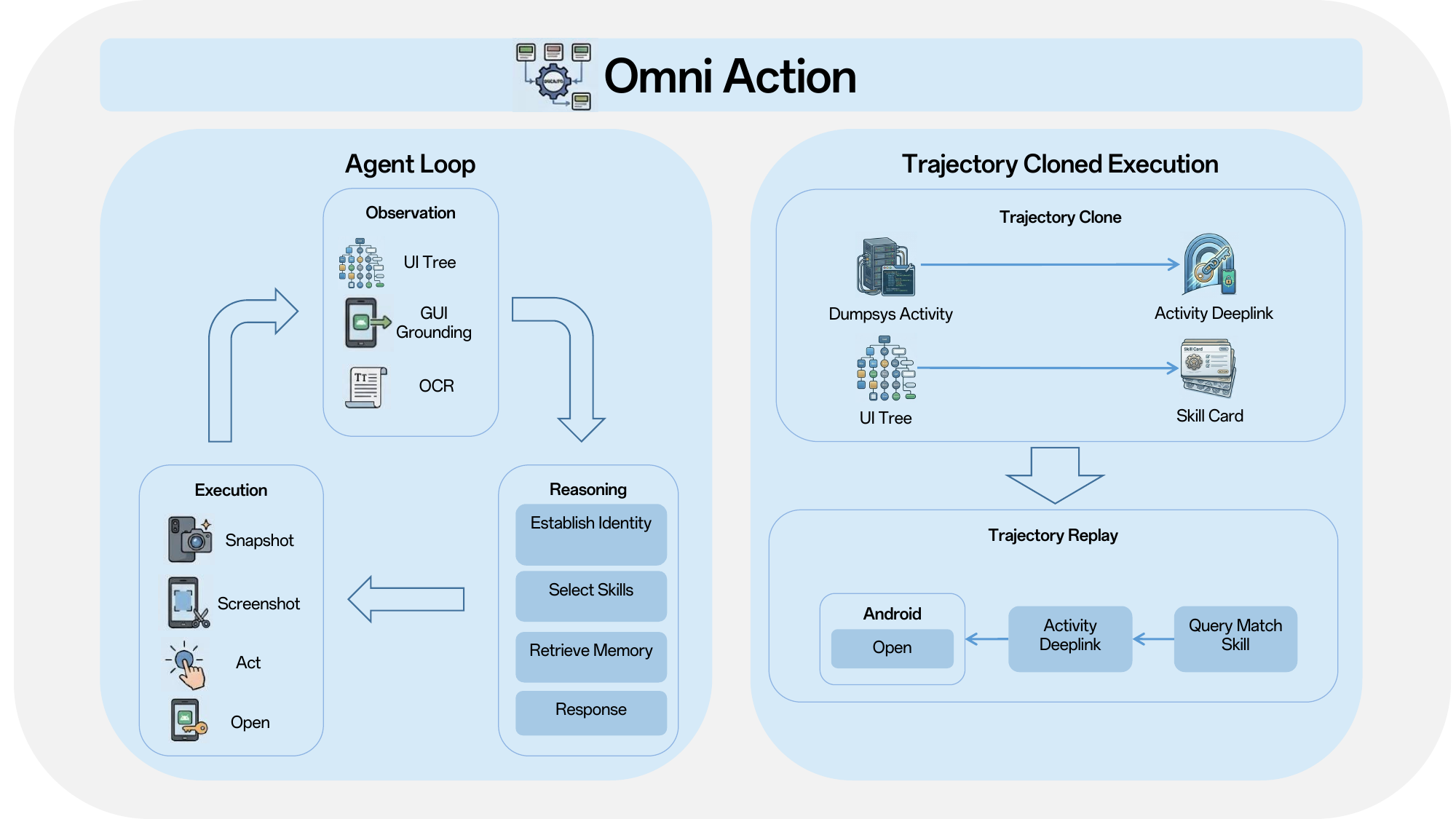}
  \caption{Overview of Omni Action in the app ecosystem: agent loop and trajectory-cloned execution.}
  \label{fig:omni-action-ecosystem-overview}
\end{figure}

\subsection{Omni Action as Trajectory Cloned Execution}
X-OmniClaw further extends action from one-shot execution to trajectory understanding and cloned execution. This shift matters in practice because mobile execution must avoid erroneous clicks, shorten long action chains, and remain robust to interruptions such as advertisements or unstable intermediate pages.

\paragraph{Behavior Cloning.}
To turn real user behavior into reusable execution knowledge, X-OmniClaw records the observable interaction process at the UI layer so that the cloned behavior can be summarized as a named skill. At this stage, the goal is to capture the purpose of the behavior rather than to reproduce each action literally, for example, ``find the reward-claim entry'' or ``jump directly to a specific video-editing template.'' X-OmniClaw combines UI-state tracking, structural parsing, and multimodal visual understanding to interpret the user's interaction trajectory and extract the semantic intent behind the cloned skill.

To achieve efficient execution, we extract \texttt{deeplink} and \texttt{intent} parameters via \texttt{dumpsys activity} introspection to bypass redundant UI replays. The technical route integrates UI-tree parsing for path capture with a two-stage fallback strategy for robust entry recovery. The system first uses incremental keyword-based filtering to rapidly locate the target activity. If that fails, it falls back to full \texttt{dumpsys} parsing to ensure completeness. Finally, this workflow distills these interactions into reusable skill cards, enabling direct jumps to target states in future tasks.

\paragraph{Trajectory Replay.}

Upon matching a skill, we recover the executable ``address'' of the target page so that later invocations can jump to it. To avoid execution failures caused by dynamic UI changes, we bypass the original click-by-click path. This allows us to maintain precise control even over non-standard application pages.

In practice, we have already instantiated a set of directly replayable or fast-entry routes across four major categories---e-commerce, local services, short-video platforms, and search---enabling one-click access to target tasks. Even when a query does not match a previously cloned end-to-end skill, X-OmniClaw can still execute rapid actions through the same deeplink-based techniques. For pre-instantiated scenarios, intent localization decomposes the request into a triple $\langle$\emph{target app, action type, parameter slots}$\rangle$ and maps the result to an application-native entry point, enabling fast access without requiring a fully cloned skill. Concrete examples are presented in the demo scenarios below.

At the current stage, this replay pipeline does not directly depend on step-by-step reproduction of the user's original actions. Nevertheless, the system still preserves the corresponding execution traces. We view these records as a reserved capability for future development: in more complex scenarios, such as cases with larger interface drift, missing stable direct-entry routes, or a need to recover finer-grained interaction semantics, trace-level replay can serve as a complementary mechanism alongside executable descriptors rather than replacing them.

%% file: sec/6_use_cases_and_demo_scenarios.tex
\section{Use Cases and Demo Scenarios}
\label{sec:use_cases_demo}

We sketch three demo tracks: \textbf{Real-world Copilot Assistant} (Sec.~6.1), \textbf{Proactive Personalized Services} (Sec.~6.2), and \textbf{Behavior Cloning and Trajectory Replay} (Sec.~6.3).

%% file: subsec/6_1.tex
\subsection{Scenario A: Real-world Copilot Assistant}

\paragraph{Demo A1: Camera-informed execution.}
The user points the camera at a real-world object and asks a situated question such as ``check the price of this product.'' Conditioned on live camera observations together with the spoken query, the agent performs multimodal perception to identify salient product cues, infer the shopping intent, and resolve the target destination. Specifically, the query is first perceived as a semantic instruction, following the same scene-grounded intent understanding described in Section~\ref{sec:cross-world-perception}. After that, the resulting intent is decomposed into an (app, keywords) pair. From there, X-OmniClaw resolves the localized intent into the target application, the required action, and the relevant parameters, and then maps them to an application-native entry point through the deeplink-based fast-entry mechanism described in Section~\ref{sec:skill-clone}, enabling one-click access to the target task.
\begin{figure}[htbp]
  \centering
  \begin{subfigure}[t]{0.48\linewidth}
    \centering
    \includegraphics[width=\linewidth]{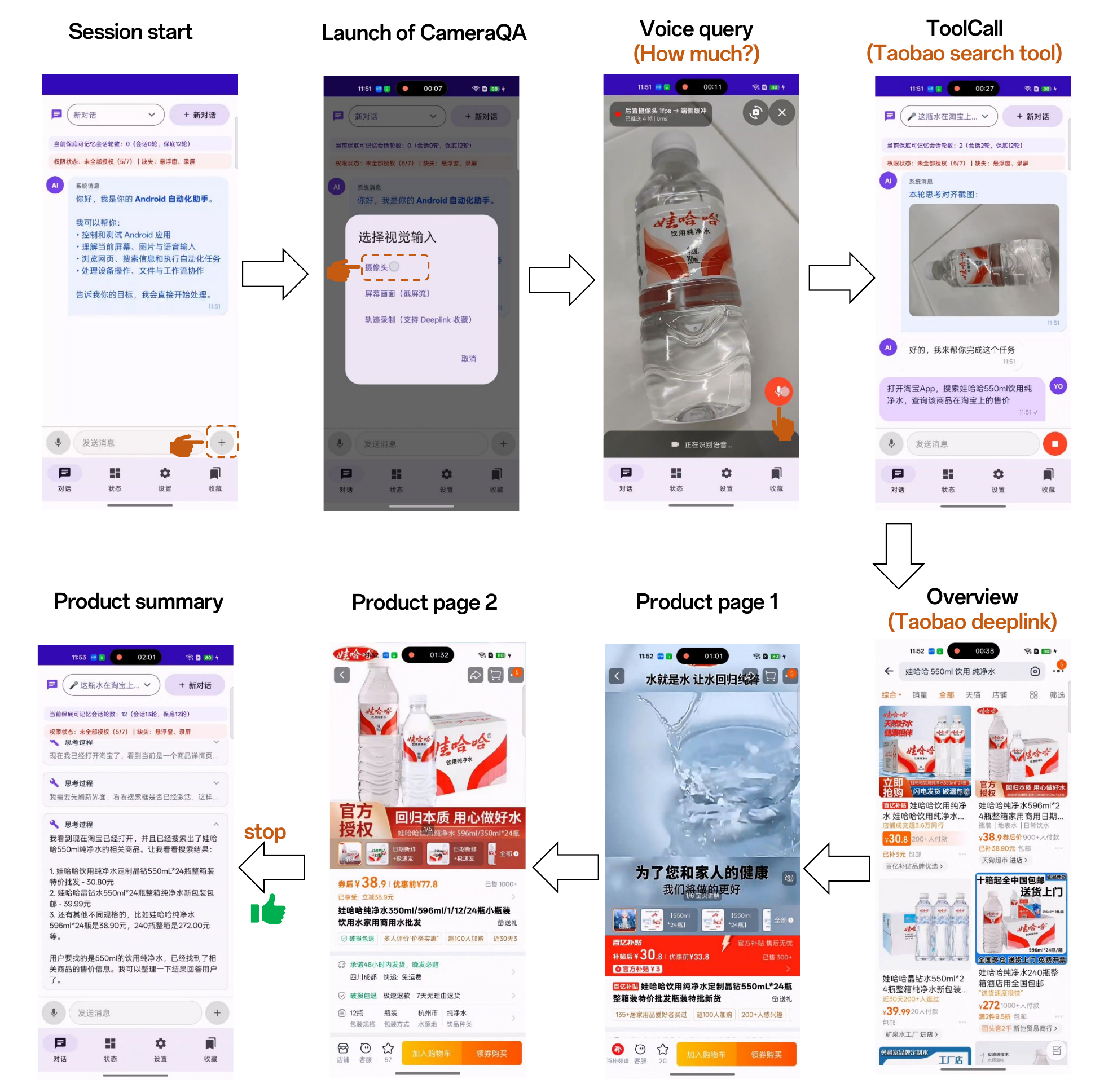}
    \caption{}
    \label{fig:demo-a1-taobao-scroll-screenshot-extract}
  \end{subfigure}\hfill
  \begin{subfigure}[t]{0.48\linewidth}
    \centering
    \includegraphics[width=\linewidth]{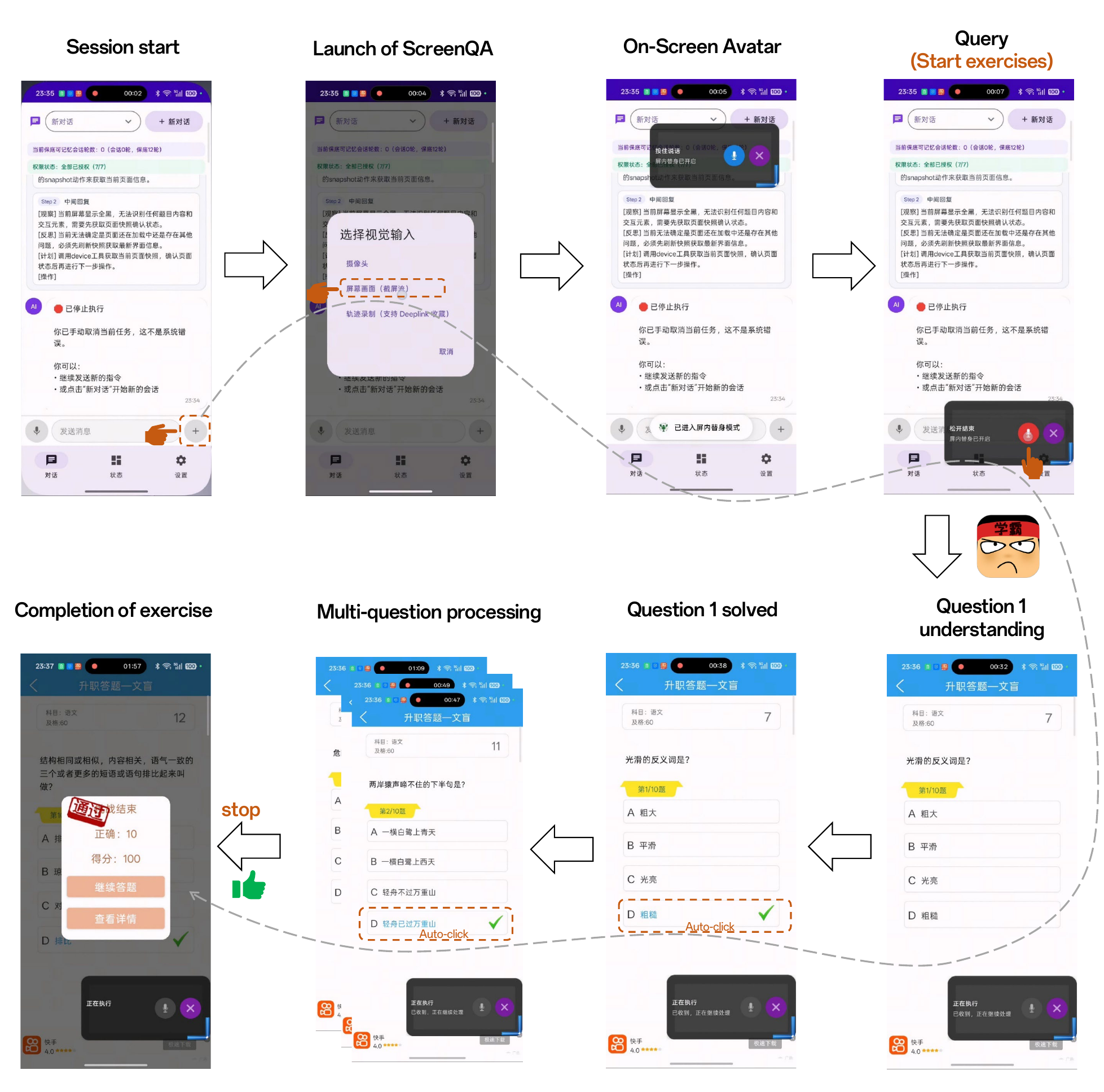}
    \caption{}
    \label{fig:demo-a2-autoproblemsolve}
  \end{subfigure}
  \caption{Scenario~A illustrations: camera-informed execution with direct app entry and result extraction (a); multi-step execution driven by screen-projected multimodal context (b).}
  \label{fig:demo-a1-a2-side-by-side}
\end{figure}

As illustrated in Figure~\ref{fig:demo-a1-taobao-scroll-screenshot-extract}, after the target app opens, the agent enters a scroll--screenshot--extract loop rather than leaving the user to browse results manually. It scrolls the result list several times, captures a screenshot after each pass, and applies VLM-based reading to extract domain-specific structured fields---for example, prices and sales in e-commerce interfaces, or ratings and distances in local-service listings. These records are materialized as structured session artifacts rather than left as transient observations, which makes follow-up interaction and later reuse more stable. The extracted evidence is then condensed into a concise summary citing concrete numbers, and a follow-up utterance such as ``open the second item'' can continue the session without re-grounding.
\paragraph{Demo A2: ScreenAvatar execution.}

When the task is specified by screen projections rather than the camera view, X-OmniClaw acts as a \textbf{ScreenAvatar}---a digital surrogate that resides on the display to execute tasks on the user's behalf. We deploy a lightweight floating companion that follows the active interface and initiates execution upon a microphone trigger. Figure~\ref{fig:demo-a2-autoproblemsolve} illustrates this end-to-end process for complex, long-chain requests such as ``help me consecutively solve these problems.'' To fulfill such requests, we combine real-time screen content with spoken intent to form a multimodal understanding of the task state. We then plan and drive a long-horizon sequence of actions across the foreground interface. By continuously interpreting intermediate UI states and updating the execution strategy, we achieve autonomous multi-step execution grounded in live-screen context. This ensures that the \textbf{ScreenAvatar} stays perfectly aligned with the foreground application throughout the entire multi-step completion process, requiring minimal user intervention.

%% file: subsec/6_2.tex
\providecommand{\CapCutApp}{\textbf{CapCut}}
\begin{figure}[htbp]
  \centering
  \includegraphics[width=\linewidth]{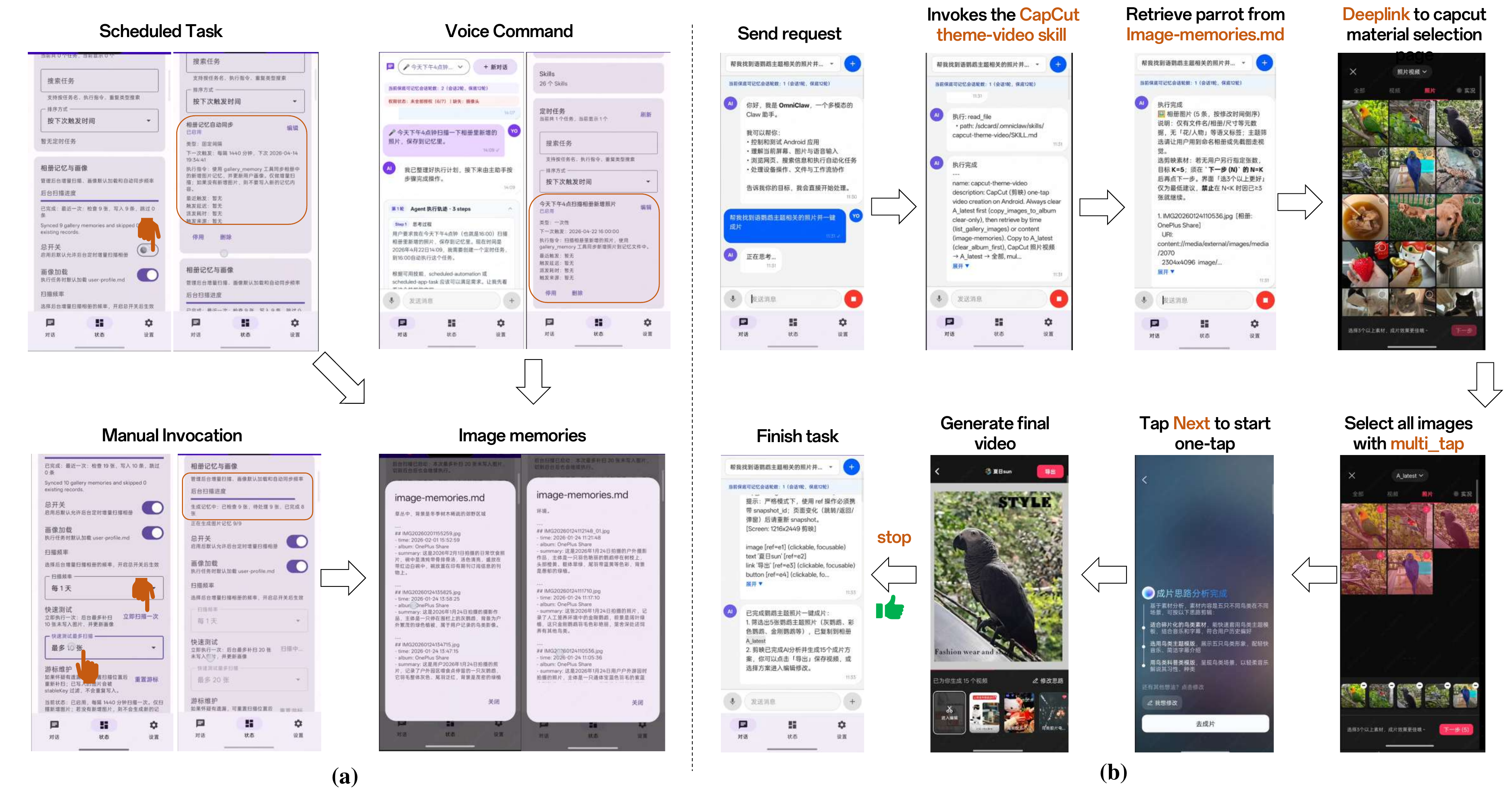}
  \caption{Illustration of the theme-based one-tap video composition: (a) multimodal gallery memory and (b) theme-based {\CapCutApp} automation (Demo~B).}
  \label{fig:demo-B-capcut-theme-video}
\end{figure}

\subsection{Scenario B: Proactive Personalized Services}

\paragraph{Demo B: Memory-based One-Tap Video.}
After a trip, manually organizing theme-related photos and turning them into a short video is often tedious. X-OmniClaw addresses this scenario through an end-to-end workflow that couples multimodal perception (Section~\ref{sec:cross-world-perception}), long-term memory (Section~\ref{sec:multimodal-long-term-memory}), and deeplink-based action (Section~\ref{sec:skill-clone}). Its proactive aspect lies in scheduled idle-time photo archiving: before the user issues an editing request, the system can periodically organize recent photos and convert them into semantic memory. As summarized in Figure~\ref{fig:demo-B-capcut-theme-video}, the system first builds semantic memory for the user's photos during idle time, then retrieves theme-matched assets when a video request arrives, and finally drives the target editing app through real on-screen interactions.


\textbf{Multimodal memory retrieval.}
The \texttt{gallery\_memory} skill incrementally scans photos and appends structured semantic entries to a \texttt{Markdown} memory file. This memory generation process can be triggered proactively by scheduled idle-time tasks, or explicitly through voice commands and manual invocation, all converging into a unified long-term memory. Later, when users issue high-level semantic requests such as ``find all parrot-themed photos and generate a highlight album in one click'', the system retrieves filenames consistent with the query, reconciles them with the media provider, and consolidates the selected assets into a task-isolated staging folder so that later automation can operate on a compact set instead of the full gallery.


\textbf{One-tap video generation.}
Under a conventional workflow, the user must first locate the correct editing entry, browse through a large gallery, and select the desired assets one by one, a process that often takes a few minutes or longer. X-OmniClaw turns this manual routine into a short automated path by reusing distilled action skills. It leverages a \texttt{deeplink} to jump directly into {\CapCutApp}'s \texttt{one-tap video composition} surface, uses UI snapshots to locate the staging folder prepared above, and applies \texttt{multi\_tap} actions to batch-select the chosen photos. This compresses the workflow into a few automated steps and substantially reduces manual intervention.

%% file: subsec/6_3.tex
\begin{figure}[htbp]
    \centering
    \includegraphics[width=\linewidth]{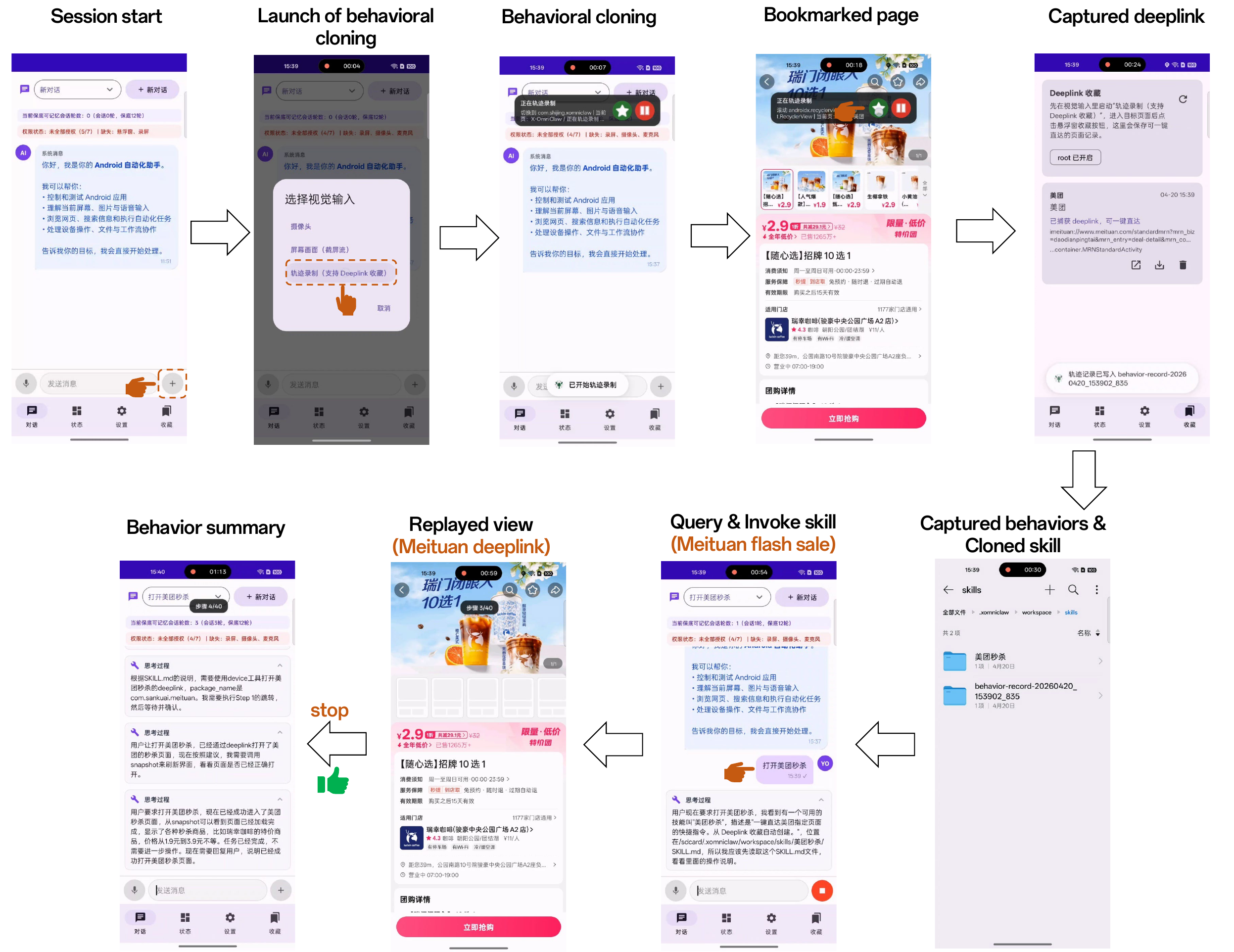}
    \caption{Illustration of instant portal to a Meituan flash-sale page (Demo~C).}
    \label{fig:demo-c-behavior-clone}
  \end{figure}
\subsection{Scenario C: Behavior Cloning and Trajectory Replay}

\paragraph{Demo C: Instant Portal to a Meituan Flash-Sale Page.}

This demo makes the trajectory cloned execution in Section~\ref{sec:skill-clone} concrete. In many apps, useful task-specific entry points are buried under several pages; X-OmniClaw therefore allows the user to clone the navigation path once, remember it as a reusable skill, and invoke it again with a single query in later sessions. As shown in Figure~\ref{fig:demo-c-behavior-clone}, the system first observes navigation at the UI layer, then converts it into a persistent, replayable artifact.


\textbf{Behavior Cloning.} When the user triggers a ``behavior cloning'' from the persistent tracking overlay, the system captures the current Activity's full launch Intent---including action, data URI, and extras---via system-level introspection. The captured entry is then saved together with a lightweight page summary as a structured bookmark and materialized as a generated skill that can be invoked again later. The captured bookmarks and generated skills serve as navigation primitives: higher-level agent plans can call them directly without knowing each app's internal URI scheme.

\textbf{Trajectory Replay.} At replay time, the user can invoke the generated skill through a natural-language query, after which the system performs query and skill processing before launching the bookmarked page. A multi-tier launch strategy is then applied: the system first attempts a full Intent or deeplink replay with all captured parameters; if that fails (\eg due to unexported Activities), it falls back to progressively simpler launch methods, down to task-stack restoration that brings the app's last-viewed page to the foreground. This graduated approach enables precise page restoration even for applications that expose no public Deep Links, covering scenarios such as ``return to the exact video'' or ``reopen a specific product detail.''


%% file: sec/7_conclusion.tex
\section{Conclusion and Future Work}
\label{sec:conclusion}

This report presented \textbf{X-OmniClaw}, an edge-native omni-modal mobile agent for Android that treats the smartphone as a unified substrate for perception, memory, and action. Across the system design, we argued that mobile agency should not be reduced to isolated screenshot-based automation or cloud-hosted remote control. Instead, the phone itself can serve as a first-person computational interface that continuously integrates on-screen UI state, real-world context, speech input, and personalized history into a single executable loop. Building on this view, \textbf{Omni Perception} provides unified ingress and scene-grounded intent understanding, \textbf{Omni Memory} maintains runtime continuity while distilling multimodal device-resident data into persistent personal knowledge, and \textbf{Omni Action} closes the loop by mapping these signals to robust execution, leveraging hybrid UI understanding and behavior cloning to transform high-level goals into reusable, executable skills. The demo scenarios further show how these components come together in practice, enabling real-world copilot assistance, proactive personalized services, and trajectory-cloned execution. 

Looking ahead, the evolution of \textbf{X-OmniClaw} focuses on three strategic pillars to further enhance system intelligence and efficiency. First, we aim to incorporate a \textbf{self-evolving mechanism} that iteratively refines execution trajectories, distilling complex reasoning chains into compact representations to minimize token consumption and response latency. Second, the architecture is transitioning toward \textbf{dynamic memory evolution}, implementing semantic consolidation and selective forgetting to ensure the user profile remains relevant and high-quality over time. Finally, we are advancing a \textbf{device-cloud synergy} that prioritizes the privacy-preserving and lightweight advantages of on-device processing for daily tasks, while selectively offloading intensive open-domain reasoning to cloud-based LLMs via secure, intent-aware gateways. Together, these advancements ensure a more resource-efficient, private, and continuously improving intelligent agent experience.

This work is conducted strictly for academic research. The proposed framework aims to explore the upper bounds of mobile agents in complex environments and does not target any specific commercial applications or platforms. To support open research and community-driven development, we will release all of our code, assets, and related artifacts as open source, and we will continue to update the project as the system evolves.